\documentclass[runningheads]{llncs}

\usepackage{graphicx}
\usepackage{comment}
\usepackage{amsmath,amssymb}
\usepackage{color}
\usepackage{url}
\usepackage{hyperref}
\usepackage{bbm} 
\usepackage{subfigure}

\newif\ifreview
\reviewfalse

\ifreview
	\usepackage{lineno}

	\linenumbers
\fi

\begin{document}


\def\SubNumber{87}

\def\GCPRTrack{Regular Track}

\title{ScaleNet: An Unsupervised Representation Learning Method for Limited Information}

\ifreview
\else

	\author{Huili Huang\inst{1}\orcidID{0000-0002-0183-8273} \and
	M. Mahdi Roozbahani\inst{2}\orcidID{0000-0001-5462-1409} }
	
	\authorrunning{H. Huang et al.}
	
	\institute{
    School of Computational Science and Engineering, 
    Georgia Institute of Technology
    756 W Peachtree St NW, Atlanta, GA 30308, USA\\
    \email{hhuang413@gatech.edu}
    \and
    School of Computational Science and Engineering, 
    Georgia Institute of Technology
    756 W Peachtree St NW, Atlanta, GA 30308, USA\\
    \email{mahdir@gatech.edu}}
\fi
\maketitle              

\begin{abstract}
Although large-scale labeled data are essential for deep convolutional neural networks (ConvNets) to learn high-level semantic visual representations, it is time-consuming and impractical to collect and annotate large-scale datasets. A simple and efficient unsupervised representation learning method named ScaleNet based on multi-scale images is proposed in this study to enhance the performance of ConvNets when limited information is available. The input images are first resized to a smaller size and fed to the ConvNet to recognize the rotation degree. Next, the ConvNet learns the rotation-prediction task for the original size images based on the parameters transferred from the previous model. The CIFAR-10 and ImageNet datasets are examined on different architectures such as AlexNet and ResNet50 in this study. The current study demonstrates that specific image features, such as Harris corner information, play a critical role in the efficiency of the rotation-prediction task. The ScaleNet supersedes the RotNet by $\approx 7\%$ in the limited CIFAR-10 dataset. The transferred parameters from a ScaleNet model with limited data improve the ImageNet Classification task by about $6\%$ compared to the RotNet model. This study shows the capability of the ScaleNet method to improve other cutting-edge models such as SimCLR by learning effective features for classification tasks. 
\footnote{ Paper accepted in DAGM GCPR 2021, LNCS 13024, pp. 174–188, 2021}
\end{abstract}

\section{Introduction}

Deep convolutional neural networks~\cite{ConvNet} (ConvNets) are widely used for Computer Vision tasks such as object recognition~\cite{ImageNet,Girshick_2014_CVPR,FasterR-CNN} and image classification~\cite{classification}. ConvNets generally perform better when they are trained by a massive amount of manually labeled data. A large-scale dataset allows ConvNets to capture more higher-level representations and avoid over-fitting. The prior studies show that these models produce excellent results when implemented for vision tasks, such as object detection~\cite{DBLP:journals/corr/Girshick15} and image captioning~\cite{DBLP:journals/corr/KarpathyF14}. However, collecting and labeling 
the large-scale training dataset is a very time-consuming and expensive task in fields such as neuroscience~\cite{ronneberger2015/U-NET}, medical diagnosis~\cite{rajpurkar2020appendixnet}, material science~\cite{material}, and chemistry application~\cite{thermochemistry}.

Researchers have investigated different approaches to learn effective visual representations on limited labeled data. Multiple studies employed data augmentation techniques such as scaling, rotating, cropping, and generating synthetic samples to produce more training samples~\cite{hu2017frankenstein,wu2015deep,inoue2018data}. Transfer learning methods are implemented to learn the high-level ConvNet-based representations in limited information content~\cite{yosinski2014transferable,Transferlearning_Survey}.

Self-supervised learning is a novel machine learning paradigm that has been employed in different fields, such as representation learning~\cite{bengio2013representation} and natural language processing~\cite{lan2019albert}. Self-supervised learning trains models to solve the pretext task to learn the intrinsic visual representations that are semantically meaningful for the target task without human annotation. Zhang and et al.~\cite{DBLP:journals/corr/ZhangIE16} introduced an image colorization method to train a model that colors photographs automatically. Doersch and et al.~\cite{DBLP:journals/corr/DoerschGE15} presented a pretext task that predicts the relative location of image patches. Noroozi and et al.~\cite{DBLP:journals/corr/NorooziF16} developed their semi-supervised model to learn the visual features by solving the jigsaw puzzle. MoCo~\cite{he2020momentum}, SimCLR~\cite{chen2020simple}, BYOL~\cite{grill2020bootstrap}, SIMSIAM~\cite{chen2020exploring} as contrastive learning methods, modify a two-branch network architectures to generate two different augmentations of an image and maximize the similarity outputs from two branches. Image clustering-based methods, such as DeepCluster~\cite{DBLP:journals/corr/deep_clustering}, SwAV~\cite{caron2020unsupervised}, and SeLa~\cite{asano2019self}, generate labels by an unsupervised learning method initially and then use the subsequent assignments in a supervised way to update the weights of the network. RotNet is one of the simplest self-supervised learning approaches that capture visual representation via rotation-prediction task~\cite{DBLP:journals/corr/revisitingssl}. The 2D rotational transformations (0°, 90°, 180°, 270°) of an image is recognized by training a ConvNet model during the pretext task. This method simplifies the self-supervised learning implementation and learns desirable visual features for downstream tasks. Since RotNet only affects the input images of the ConvNet, it can be combined with other architectures such as GAN~\cite{DBLP:journals/corr/rotationGAN} or be used for 3D structured datasets~\cite{DBLP:journals/corr/3Drotnet}.

The main milestone of the current research is to enhance the performance of self-supervised learning in the presence of limited information for current existing self-supervised learning models, such as the RotNet and SimCLR, as opposed to re-introducing a new architecture for these models. Recent studies propose several new architectures based on RotNet architectures. For example, Feng et al.~\cite{feng2019self} improved the rotation-prediction task by learning the split representation that contains rotation-related and unrelated features. Jenni et al.~\cite{jenni2020steering} claimed a new architecture to discriminate the transformations such as rotations, warping, and LCI to learn visual features. The current study focuses on improving self-supervised learning methods with limited information, including limited training samples, missing corner information and lacking color information. A multi-scale self-supervised learning model named ScaleNet is proposed to improve the quality of learned representation for limited data. This simple and efficient framework comprises the following three components:
resizing the original input images to a smaller size, feeding resized dataset to a ConvNet for the rotation recognition task, and training the CovNet using the larger size (e.g. original size) input dataset based on the parameters learned from the previous ConvNet model.
The ScaleNet method is trained using different architectures such as AlexNet~\cite{AlexNet} and ResNet50~\cite{ResNet}. Results show that the ScaleNet outperforms the RotNet by 1.23\% in the absence of image corner information using the CIFAR-10 dataset and 7.03\% with a limited CIFAR-10 dataset. The performance of the SimCLR with a limited dataset and small batch size is improved by $\sim$4\% using a multi-scale SimCLR model. The experiments outlined in this study demonstrate that the performance of the RotNet model for a larger dataset is enhanced by using the parameters learned from the ScaleNet model trained with a smaller dataset.

\section{ScaleNet}
\subsection{Self-supervised Learning based on Geometrical Transformation} \label{3.1}
	
Assume $y_i\in{Y}$ is the human-annotated label for image $x_i\in{X}$, and $y_i^*\in{Y^*}$ is the pseudo label generated automatically by a self-supervised ConvNet. Instead of training the ConvNet model $F(.)$ by human-labeled data ($X$ and $Y$) using a supervised learning method, self-supervised learning trains $F(.)$ with the images $X$ and pseudo labels $Y^*$. 
	
The length and width of images in the dataset $X$ are $R$ and $C$, respectively, and the parameter $\alpha$ is provided as the resize operator ($\alpha\leq1$). The length and width of the new dataset, $X_\alpha$, are $\alpha R$ and $\alpha C$($X_\alpha$ represents the rescaled dataset from $X$ and $x_{\alpha i}$ is one sample from dataset $X_\alpha$). 

A geometrical transformation operator is defined as $G=\left\{g(.|y)\right\}^K_{y=1}$, where $K$ denotes different geometrical transformations such as resizing or rotation. $g(X|y)$  is the geometric transformation that applies to a dataset $X$. For example, $X_\alpha^{y^*}=g(X_\alpha|y^*)$ is the resized dataset $X_\alpha$, transformed by the geometric transformation and labeled as $y^*$ by ConvNet $F_\alpha(.)$. The ConvNet model $F_\alpha(.)$ is trained on the pseudo-label dataset $X_\alpha^{y^*}$. The probability distribution over all possible geometric transformations is:
	\begin{equation}
	\begin{aligned}
	F_\alpha(X_\alpha^{y^*}|\theta) = \{F_\alpha^y(X_\alpha^{y^*}|\theta)\}^K_{y=1}
	\end{aligned}
	\end{equation}
where $F_\alpha(.)$ gets input from $X_\alpha$, $F_\alpha^y(X_\alpha^{y^*}|\theta)$ is the predicted probability of the geometric transformation with label $y$, and $\theta$ is the parameter that learned by model $F_\alpha(.)$. 
	
A cross-entropy loss is implemented in this study. Given a set of N training images $D = \left\{x_i\right\}_{i=0}^N$ and the ConvNet $F_1(.)$ as an example, the overall training loss is defined as:
	\begin{equation}
	\begin{aligned}
	    Loss(D) = \mathop{min}\limits_{\theta}\frac{1}{N}\sum\limits_{i=1}^Nloss(x_i,\theta)
	\end{aligned}
	\end{equation}
where the $loss(x_i,\theta)$ is the loss between the predicted probability distribution over K and the rotation $y$:
	
    \begin{equation}
	\begin{aligned}
	    loss(x_i,\theta) = -\frac{1}{K}\sum_{y=1}^K\log(F^y_1(g(x_{i}|y)|\theta))
	\end{aligned}
	\end{equation}
	
	\begin{figure*}[t]
	\begin{center}
	\includegraphics[height=8cm]{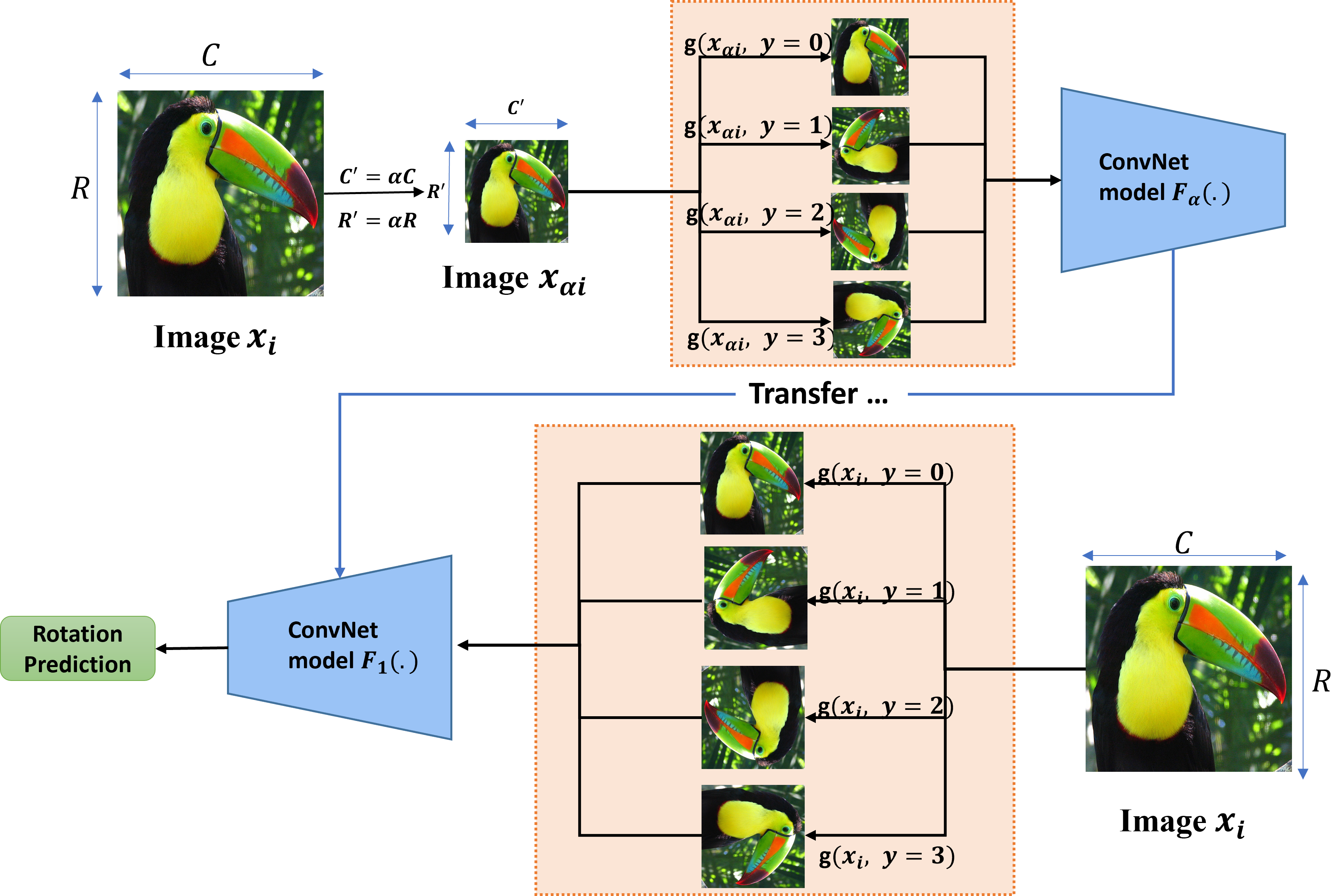}
    \end{center}
		\caption{
		The Input image $x_i$ is first scaled to a smaller image $x_{\alpha i}$ and rotated 0, 90, 180, 270 degrees. The ConvNet model $F_\alpha(.) $ is trained to predict the rotation degree of an image $x_{\alpha i}$. The ConvNet model  $F_\alpha(.)$ is then transferred to the ConvNet model $F_1(.)$ to train the rotation-prediction task. 
		}
		\label{fig:example1}
	\end{figure*}

\subsection{Multi-Scale Rotation-Prediction based on Different Scales}\label{2.2}

A geometrical transformation $G$ is defined to rotate images by a multiplier of 90 degrees~\cite{DBLP:journals/corr/rotation}. $G = \left\{g(X|y)\right\}^4_{y=1}$ represents the rotation and  $y = 0, 1, 2, 3$ is the multiplier to generate 0, 90, 180, 270 rotated images, respectively. The horizontal and vertical flipping operations are implemented to rotate images. The rotation task is performed in three ways: transposing and then vertical flipping to rotate an image by 90 degrees, vertical flipping and then transposing to rotate by 270 degrees, and vertical flipping and then horizontal flipping to rotate by 180 degrees. 

The main goal of the rotation-prediction pretext task is to enforce the ConvNet to recognize high-level objects such as the horse's leg or bird's eyes~\cite{DBLP:journals/corr/rotation,jing2018self}. Gidaris et al.~\cite{DBLP:journals/corr/rotation} demonstrated that the RotNet model focused on almost the same image regions as the supervised model by comparing the attention maps. 
Similar to other data augmentation methods such as flipping and cropping~\cite{8243510dataaugmentation}, rotating images hinder the model from learning impractical low-level visual features that are trivial information for downstream tasks~\cite{DBLP:journals/corr/rotation}. The RotNet learns these features by unlocking different underlying structures of images such as texture, shape, and high-level semantic features by rotating images.  

The multi-scale rotation prediction is considered a pretext task (Fig.~\ref{fig:example1} ). The original images are contracted using the resize operator $\alpha$. Each image is transformed counterclockwise to reproduce four images: no rotation, rotation by 90, rotation by 180, and rotation by 270 degrees and  labeled as 0, 1, 2, 3, respectively. The contracted images are then fed to a ConvNet model $F_\alpha(.)$ for a rotation-prediction task. After that, the parameters learned from $F_\alpha(.)$ are transferred to another larger scale model. For example, The training steps for a two-model ScaleNet trained with the resize operator combination $\alpha = [0.5, 1]$ are:
\begin{itemize}
\item Obtaining dataset $X_{0.5}$ using the resize operator $\alpha = 0.5$.
\item Feeding the dataset $X_{0.5}$ to ConvNet $F_{0.5}(.)$ to learn the rotation-prediction task.
\item Training CovNet $F_1(.)$ using the original input dataset $X$ based on the parameters learned from model $F_{0.5}(.)$. 
\end{itemize}
Note that the ScaleNet allows users to train more than one pre-trained model before training the $F_1(.)$. For example, the three-model ScaleNet is trained with the resize operator combination $\alpha = [0.5, 0.75, 1]$.

\section{Experiments}
 The rotation-prediction task is trained on different scales of images. Note that the current study focuses on enhancing the self-supervised learning method with limited information. First, the value of the resize operators $\alpha$ and the learning rate are analyzed to find the best model for the ScaleNet. Second, the effects of limited information are evaluated. For example, missing corner information and lacking color information on the pretext task is studied. Third, a multi-scale self-supervised learning technique is implemented in SimCLR with a small batch size and limited information to confirm the effectiveness of the ScaleNet further. Finally, the relationship between the quality of the feature maps and the depth of the model is analyzed. All experiments are trained on the CIFAR-10~\cite{CIFAR10} or ImageNet~\cite{ImageNet} datasets in this study. Input data are standardized (zero mean and unit standard deviation). A Stochastic gradient descent (SGD) is applied to optimize the rotation-prediction network, and Adaptive Moment Estimation (Adam) is applied to train the SimCLR network. NVIDIA K80 is used for ImageNet and SimCLR models and RTX 2070 super for other models. 

\subsection{CIFAR-10 Experiment}\label{4.1EX-CIFAR10}
\subsubsection{Determining the Resize Operator:}\label{resize-operator}
\setlength{\tabcolsep}{4pt}
The resize operator $\alpha$ is evaluated quantitatively using different multi-model ScaleNets. Three kinds of models are trained in this Section: RotNet model, two-model ScaleNet, three-model ScaleNet. The RotNet model is similar to the $\alpha=1$ model, where the rotation recognition task is performed on original images. The two-model ScaleNet is initially trained on a smaller size of images using one of the resize operators of $\alpha=0.25$, $\alpha = 0.5$, or $\alpha=0.75$. The pre-trained model's parameters are then transferred to train a new model with an $\alpha$ value of 1, which is the pre-trained model for the downstream task. For the three-model ScaleNet, model's parameters with a smaller $\alpha$ value (e.g., $\alpha$=0.5) are used to train the model with a larger $\alpha$ value (e.g., $\alpha=0.75$). The weights of the conv. layers for the downstream task model are transferred from the last pre-trained ScaleNet model (e.g., $\alpha=0.75$). Since the same dataset is trained, there is no need to freeze the weights of the conv. layers~\cite{soekhoe2016impact}. The initial learning rate of the CIFAR-10 classification task (downstream task) is 0.001, which decays by a factor of 5 when the epoch reaches 80, 160, 200. All experiments are constructed using ResNet50 architecture with the batch size 128, momentum value 0.9, and the weight decay value $5e-4$. Each ConvNet model is trained with 100 epochs.
\begin{table*}[b]
\begin{center}
\begin{tabular}{ccccc}
\hline\noalign{\smallskip}
		Method &$\alpha$ &Learning rate &Pretext task & classification\\
		\noalign{\smallskip}
		\hline
		\noalign{\smallskip}
		Two-model ScaleNet &{0.5,1}& 0.1, 0.1& 84.73 & 86.81\\
		Two-model ScaleNet & {0.5,1}& 0.1, 0.05 & 84.65&\textbf{88.19}\\
		Two-model ScaleNet & {0.5,1}& 0.1, 0.01 & 83.0& 88.00\\
		Three-model ScaleNet & {0.5,0.75,1} & 0.1, 0.05,0.05 & 85.77& 86.16\\
		Three-model ScaleNet & {0.5,0.75,1} & 0.1, 0.05,0.01 &85.08&\textbf{88.04}\\
		\hline
    \end{tabular}
\end{center}
\caption{Evaluation of different learning rates for the ScaleNet model. All models are trained on ResNet50. Different Learning rate combinations are tested based on the two-model and three-model ScaleNet. $\alpha=0.5$ corresponds to the 16 $\times$ 16 input images. $\alpha=0.75$ corresponds to 24 $\times$ 24 input images. }
\label{table:EX_LR}
\end{table*}
 A smaller learning rate is used to train a larger $\alpha$ value model in the ScaleNet to prevents the significant distortion of the ConvNet weights and avoids over-fitting~\cite{ng2015deep}. The different learning rate combinations are investigated to confirm the study by Ng et al.~\cite{ng2015deep}. For example, $[0.1,0.1]$, $[0.1,0.05]$, and $[0.1,0.01]$ are selected for the two-model ScaleNet. The initial learning rate decays by a factor of 5 when the epoch reaches 30, 60, and 80. Since there are ten categories in the CIFAR-10 dataset, the classification layer output is adjusted from four (four rotation degrees) to ten for the downstream task. Table~\ref{table:EX_LR} shows that setting a lower learning rate improves the performance of the downstream task and the learning rate combinations of $[0.1,0.05]$, and $[0.1,0.05,0.01]$ achieve highest accuracy in the classification task. The learning rate combination $[0.1,0.05]$ for two-model ScaleNet and $[0.1,0.05,0.01]$ for three-model ScaleNet are then used in the evaluation of the resize operator $\alpha$ in Table~\ref{resize-operator}.

\setlength{\tabcolsep}{4pt}
\begin{table*}[t]
\begin{center}
\begin{tabular}{ccccc}
		\hline\noalign{\smallskip}
		Method &$\alpha$ &Learning rate &Pretext task & classification\\
		\noalign{\smallskip}
		\hline
		\noalign{\smallskip}
		RotNet  & 1 & 0.1 & 83.64 & $87.17 \pm 0.13$\\
		Two-model ScaleNet & {0.25,1}& 0.1, 0.05& 82.71 & $86.19 \pm 0.16$\\
		Two-model ScaleNet & {0.5,1}& 0.1, 0.05 & 84.65&\textbf{$88.19 \pm 0.04$}\\
		Two-model ScaleNet & {0.75,1}& 0.1, 0.05 & 85.13& $87.63 \pm 0.21$\\
		Three-model ScaleNet & {0.25,0.5,1} & 0.1, 0.05,0.01 & 81.49& $86.31 \pm 0.17$ \\
		Three-model ScaleNet & {0.25,0.75,1} & 0.1, 0.05,0.01 & 83.14& $86.59 \pm 0.35$ \\
		Three-model ScaleNet & {0.5,0.75,1} & 0.1, 0.05,0.01 & 85.08& $88.00 \pm 0.05$\\
		\hline
\end{tabular}
\end{center}
\caption{Evaluation of the resize operator $\alpha$. All models are trained on ResNet50. $\alpha=0.25$ corresponds to the 8 $\times$ 8 input images. $\alpha=0.5$ corresponds to 16 $\times$ 16 input images. $\alpha=0.75$ corresponds to 24 $\times$ 24 input images. Each experiment was run three times. The same learning rate 0.1 is implemented for the RotNet~\cite{DBLP:journals/corr/rotation}}
\label{table:EX_RESCALE_OPERATOR}
\end{table*}

Table~\ref{table:EX_RESCALE_OPERATOR} shows that the ScaleNet model with an $\alpha$ value of 0.5 outperforms the RotNet by about 1\%. Although the improvement is minimal, it manifests itself when limited information is available, explained in Section~\ref{limited-data} latter. As shown in Table~\ref{table:EX_RESCALE_OPERATOR}, the $\alpha=0.5$ operator performs better than other operators. Suppose the pixel is identified by a pair of coordinates $(x,y)$, only pixels in even coordinates like $(2,2)$ are retained during the bi-linear interpolation with $\alpha=0.5$. Other resize operators compute the resized pixel value by averaging over four surrounding pixels. The $\alpha=0.25$ scales down images to a tiny size ($8\times8$) that obscures essential information for ConvNets tasks. Considering Table~\ref{table:EX_LR} and Table~\ref{table:EX_RESCALE_OPERATOR} results, the two-model ScaleNet with an $\alpha$ value of $[0.5,1]$, and a learning rate with the combination of $[0.1,0.05]$ are employed for later experiments in this study on the CIFAR-10 dataset. 

\subsubsection{ScaleNet performance using limited information:}\label{limited-data}
\begin{table*}
\begin{center}
\begin{tabular}{cccc}
			\hline\noalign{\smallskip}
			Method &Pretext task&classification(4K)&classification\\
			\hline
			\noalign{\smallskip}
			\noalign{\smallskip}
			RotNet & 82.45 & 71.98 & 87.17 \\
			Harris RotNet & 50.38 & 62.80 & 86.04\\
			Hybrid RotNet& 81.76 & 69.32 & 87.17 \\
			Gray-scale RotNet & 80.33 & 70.63 & 86.96\\
			Gray-scale-harris RotNet & 65.34 & 61.63 & 85.77\\
			ScaleNet & 84.65 & 73.69(+1.71\%) & 88.19(+1.02\%)\\
		    Harris ScaleNet & 45.83 & 69.83(+7.03\%)& 87.27(+1.23\%)\\
		    Hybrid ScaleNet & 81.79 & 73.16(+3.84\%) & 88.12(+0.95\%)\\
		    Gray-scale ScaleNet & 80.71 & 72.48(+1.85\%) & 87.2(+0.24\%)\\
		    Gray-scale-harris ScaleNet & 69.65 & 68.09(+6.46\%) & 86.04(+0.27\%) \\
			\hline
		\end{tabular}
\end{center}
\caption{Experiments are performed using the RotNet and ScaleNet architecture. The input of RotNet/ScaleNet, Harris RotNet/ScaleNet, Hybrid Rot/ScaleNet, Gray-scale RotNet/ScaleNet, Gray-scale-harris RotNet/ScaleNet are the original images, the images without corner information, a random combination of original and Harris images, pseudo-gray-scale images, and pseudo-gray-scale images without corner information, respectively. Each experiment runs three times to get the average results}
\label{table:EX_HARRIS}
\end{table*}

The effects of corner features, the color of images, and limited data are examined for the transformational model in Section \ref{limited-data}. The limited information is exercised in this study using a combination of the limited CIFAR-10 data (randomly selecting 4000 out of 50000), missing corner information, and gray-vs-color images. A two-model ScaleNet is trained using resize operators of $[0.5,1]$ as explained in Table~\ref{table:EX_LR} and Table~\ref{table:EX_RESCALE_OPERATOR}. Three models are constructed in this study: the model using the original/pseudo-gray-scale image, the Harris model that is trained on the original/pseudo-gray-scale images with missing corner information, and the hybrid model where the input images are a random combination of original/pseudo-gray-scale images with and without corner information. The pseudo-gray-scale images are trained based on the RotNet and ScaleNet to study the color effects of rotation-prediction results. A pseudo-gray-scale image is generated by replacing each channel with its gray-scale image to adjust the three-channel input for the pretext and downstream task. Although corners maintain only a small percentage of an image, it restores crucial information such as invariant features. The corner information of images is extracted by the Harris corner detector~\cite{harris1988combined}. The detected corner pixels by Harris are replaced with 255 (white dot), and the new images (Harris images) with missing corner information are fed to ConvNet models (Harris models).

Table~\ref{table:EX_HARRIS} shows the pretext and the classification task accuracy reduce due to the absence of the corner information for both the RotNet and ScaleNet models. The ScaleNet noticeably outperforms the RotNet by about $7\%$ for the classification task with limited information (missing corners and a limited number of data). This experiment clearly shows the adjustment of the ScaleNet to understand and detect influential features with limited information, that will be crucial for many fields with a lack of data~\cite{frid2018gan,7973178LEE,ronneberger2015/U-NET}. The same behavior is observed to a lesser extent when the gray-scale images are used instead of color images. According to Table~\ref{table:EX_HARRIS} results, a gray-scale channel would be sufficient to achieve a comparable accuracy similar to color images.

Harris corner features are invariant under rotation, but they are not invariant under scaling. A rotation-prediction task (Pretext task) is relatively dependent on corner features according to Table~\ref{table:EX_HARRIS} results. It is shown that pretext accuracy significantly reduces by removing corner features for both the ScaleNet and RotNet. The classification task for 4K data shows the light on the Harris ScaleNet improves the accuracy noticeably comparing to the RotNet. It sheds the fact that the resultant parameters by the ScaleNet are not only rotation-invariant but also scale-invariant. However, the RotNet parameters are trained to be rotation-invariant features. The ScaleNet improves the ConvNets parameters by unlocking underlying features that are crucial in classification tasks. 

\subsubsection{Multi-SimCLR performance using limited information:}
The linear evaluation based on SimCLR~\cite{chen2020simple} is introduced to further confirm the ScaleNet effectiveness in the presence of limited information. Given two different augmented data $\tilde{x_i}$ and $\tilde{x_j}$ from the same sample $x\in{X}$, the ConvNet $F(.)$ maximizes the agreement between the augmented samples. The ConvNet $F(.)$ is a symmetric architecture including the data augmentations from a family of augmentations $t\sim T$, the base encoder networks $f(.)$, and the projection heads $g(.)$. The NT-Xent~\cite{chen2020simple,oord2018representation,wu2018unsupervised} is implemented as the loss function. Given the results from the projection head $z_i = g(f(\tilde{x_i})),z_j = g(f(\tilde{x_j}))$, the loss function is:
	\begin{equation}
	\begin{aligned}
	    Loss(i,j) = -\log\frac{exp(sim(z_i,z_j))/\tau}{\sum_{k=1}^{2N}\mathbbm{1}_{[k\neq{i}]}exp(sim(z_i,z_k)/\tau)}
	\end{aligned}
	\end{equation}
where $sim(u,v)=u^\mathsf{T} v/\left \| u \right\|\left \|v \right\|$, $\tau$ is a temperature parameter, and $\mathbbm{1}_{[k\neq{i}]} \in \{0,1\}$ is an indicator function that equals 1 if $k\neq{i}$. The multi-scale SimCLR with an $\alpha$ value of $[0.5, 1]$ and a learning rate of $[0.001, 0.001]$ is selected as the best candidate model according to Table~\ref{table:EX_RESCALE_OPERATOR} and Table~\ref{table:EX_LR} experiments. 
\begin{figure*}[t]
\begin{center}
\subfigure[128 batch size]{
\includegraphics[width=5.5cm]{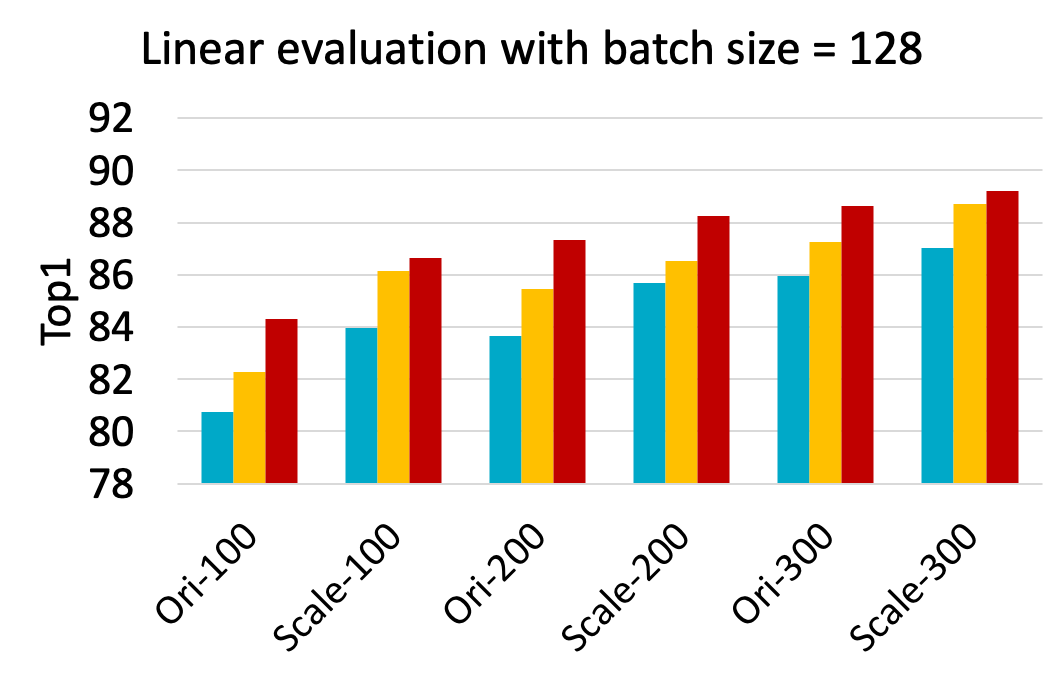}}
\subfigure[256 batch size]{
\includegraphics[width=5.5cm]{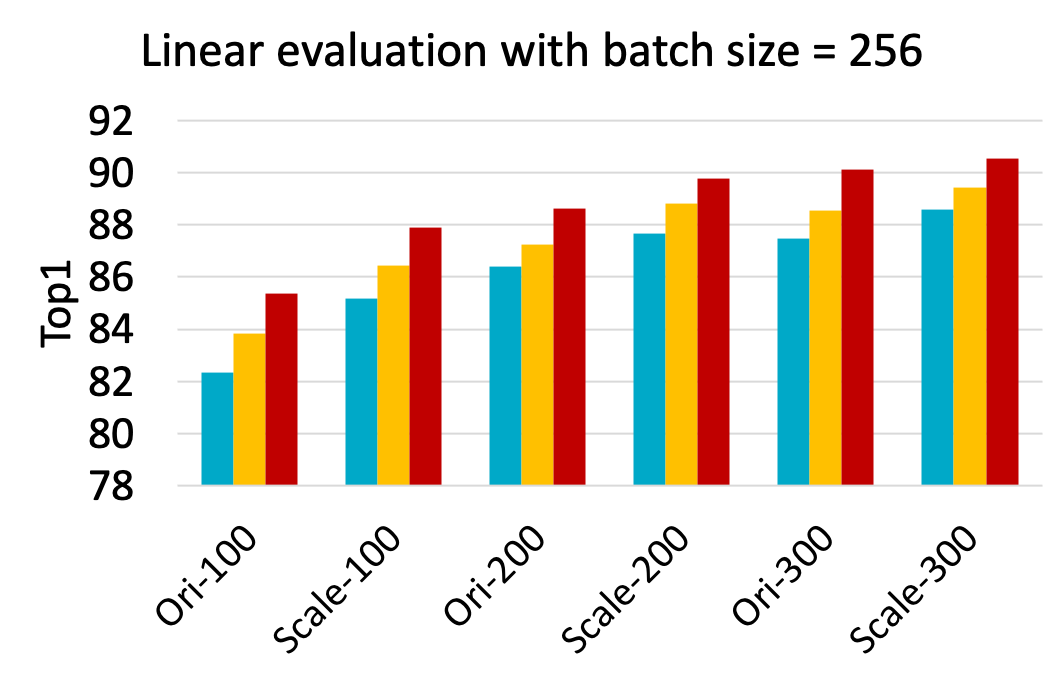}}
\subfigure[512 batch size]{
\includegraphics[width=5.5cm]{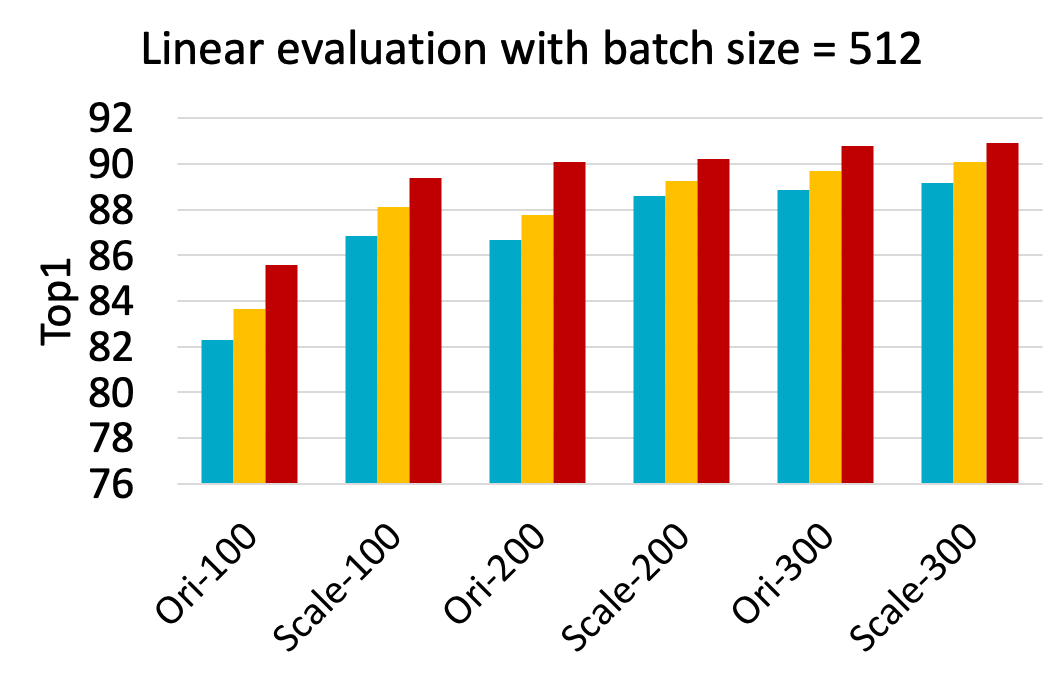}}
\end{center}
\caption{Linear evaluation of multi-scale SimCLR trained with different epochs and batch sizes on the CIFAR-10 dataset. The Ori-* and Scale-* represent the original SimCLR model~\cite{chen2020simple} and the multi-scale SimCLR, respectively. The Ori/Scale-100, Ori/Scale-200, Ori/Scale-300 show the original/multi-scale SimCLR method results trained with 100, 200, 300 epochs, respectively. The blue, yellow, and red bar show the classification task trained with 4K, 10K, 50K data, respectively. All experiments run two times to compute the average results}
\label{SimCLR_exp}
\end{figure*}
Chen's study\cite{chen2020simple} demonstrated the contrastive learning benefits from larger batch size. However, training a small-scale dataset with large batch size such as 1024 and 2048 leads to a decrease in performance, even if tuning learning rate to a heuristic~\cite{mishkin2017systematic}. The experiment in this section use small batch size to enhance the performance of SimCLR for limited data. The batch size in \{128, 256, 512\} is used in this experiment because the minimum training dataset only contains 4K samples. 0.5 and 0.001 are chosen for the temperature of the pretext  and the learning rate of the downstream task, respectively. The $F_{0.5}(.)$ model is trained with epoch size 300 for faster convergence. The trained parameters are transferred to $F_1(.)$ with \{128, 256, 512\} batch size. The Adam optimizer is applied instead of the LARS because of its effective capability to train models with small batch sizes~\cite{you2017large}. The inception crop and color distortion are used as the data augmentation methods, and ResNet50 architecture is implemented for this experiment, similar to the SimCLR study~\cite{chen2020simple}. The classification results trained for 4K, 10K, and 50K (whole dataset) data with different batch sizes are shown in Fig.~\ref{SimCLR_exp}. Note that all experiment results for the original SimCLR method are in the acceptable range of the SimCLR study~\cite{chen2020simple}. Fig.~\ref{SimCLR_exp} demonstrates the multi-scale SimCLR clearly improves the classification results with limited information. The best multi-scale SimCLR model considerably outperforms the SimCLR by 4.55\%, 4.47\%, and 3.79\% when trained on 4K, 10K, 50K data, respectively. The results further assert the ScaleNet process is an effective technique for self-supervised learning algorithm with limited information, and multi-scale training
is able to capture necessary high-level representation that cannot be learned in the existing SimCLR model\cite{chen2020simple}. 

\subsection{ImageNet Experiment}\label{4.2EX_IMAGENET}
The ScaleNet is examined by training on the ImageNet dataset similar to other self-supervised learning methods~\cite{DBLP:journals/corr/rotation,DBLP:journals/corr/NorooziF16,DBLP:journals/corr/counting}. The datasets that contain 65 and 240 samples per category are generated from ImageNet, respectively. For the ScaleNet model trained on 65000 samples, the combination of the resize operator is $[0.5, 1]$. The ScaleNet and RotNet are trained on AlexNet with batch size 192, momentum 0.9, and weight decay $5e-4$, respectively. The learning rate combination is [0.01, 0.001] for the ScaleNet to avoid over-fitting~\cite{ng2015deep}. The pretext model runs for 30 epochs with a learning rate decay factor of 10, which is applied after 10 and 20 epochs. A logistic regression model is added to the last layer for the classification task. All weights of conv. layers are frozen during the downstream task. The initial learning rate of the ImageNet classification task (downstream task) is 0.01, which decays by a factor of 5 when the epoch reaches 5, 15, 25. 
\begin{table*}[t]
\begin{center}
\begin{tabular}{ccccccc}
\hline\noalign{\smallskip}
Model& Samples &ConvB1&ConvB2&ConvB3&ConvB4&ConvB5\\
\noalign{\smallskip}
\hline
\noalign{\smallskip}
RotNet & 65,000 & 5.62 & 9.24 & 11.30 & 11.71 & 11.75\\
ScaleNet & 65,000 & \bf{5.93} & \bf{9.91} & \bf{12.05} & \bf{12.64} & \bf{12.25}\\
RotNet (from RotNet) & 240,000 & 11.30 & 17.60 & 21.26 & 22.15 & 21.90\\
RotNet (from ScaleNet) & 240,000 & \bf{14.55} & \bf{23.18} & \bf{27.75} & \bf{27.50} & \bf{26.61}\\
\hline
\end{tabular}
\end{center}
\caption{The ImageNet classification with linear layers, where the logistic regression is used in the classification layer for the downstream task. The ScaleNet and RotNet model parameters trained on 65,000 samples are used in the RotNet model with 240,000 samples as a pre-trained model and called RotNet (from RotNet) and RotNet (from ScaleNet), respectively. The rate 0.01 is implemented for the RotNet~\cite{DBLP:journals/corr/rotation}}
\label{table:EX_IMAGENET}
\end{table*}
     
 Table~\ref{table:EX_IMAGENET} shows that the generated feature maps by the ScaleNet perform better in all conv. blocks experiments comparing to the RotNet. The feature map generated by the 4th conv. layer achieves the highest classification accuracy in the ScaleNet experiment. The classification accuracy decreases after the 4th conv. layer as the ConvNet model starts learning the specific features of the rotation-prediction task. The accuracy of the ScaleNet model trained on 65,000 images is 0.93\% higher than the RotNet model.
 
 Note that the trained ScaleNet model on a limited dataset is capable of improving the RotNet model accuracy for a larger dataset when it uses the ScaleNet pre-trained parameters. This experiment is conducted by training the ScaleNet and RotNet model with 65,000 samples and transferring their parameters to train the RotNet with 240,000 samples using RotNet, respectively. Table~\ref{table:EX_IMAGENET} shows the RotNet (from ScaleNet) clearly outperforms the RotNet (from RotNet) in the classification task by 6.49\%. This result shows the capability of the ScaleNet to detect crucial features for a limited dataset and provide effective parameters from a limited dataset for a transfer learning task. 

\subsubsection{Attention map analysis of ResNet50:}\label{FILTER_VISUALIZATION}
\begin{figure}[t]
\begin{center}
\includegraphics[height=5.8cm,width=9cm]{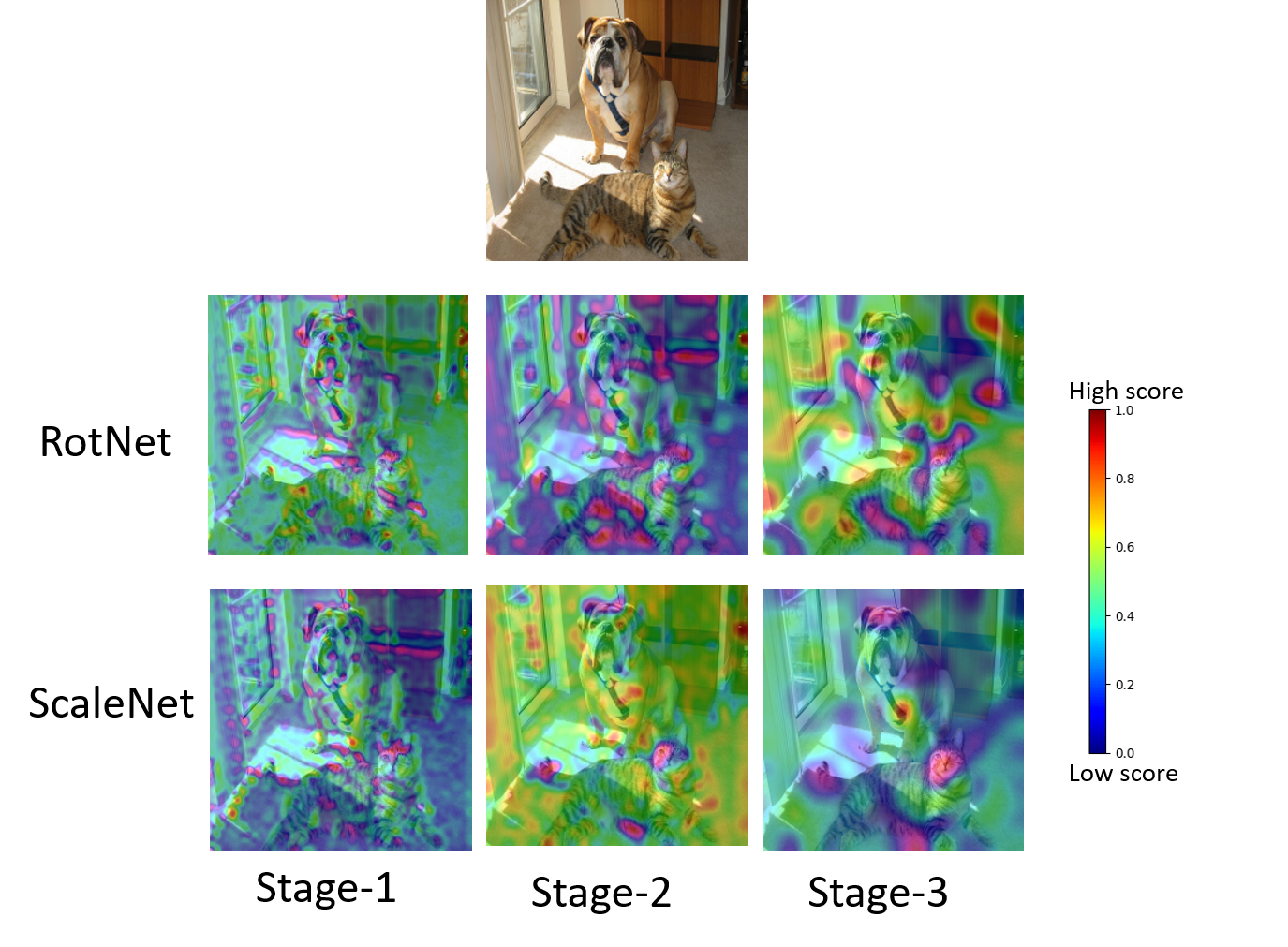}
\end{center}
\caption{ The attention maps generated by the ResNet50 model. The heat map is implemented to show the area of attention by ConvNet. Red regions correspond to a high score for the class. The first row is the original image. The second row and third row show the attention map generated by the RotNet and ScaleNet, respectively. Stage-1, Stage-2, Stage-3 are the first three stages of ResNet50, respectively. 
 	}
\label{GRADCAM}
\end{figure}

 Gidaris's study~\cite{DBLP:journals/corr/rotation} explains the critical role of the first conv. layer edge filter in the rotation-prediction task. The study proves that the RotNet catches more information in various directions than the supervised learning method. The attention maps of the ScaleNet and RotNet are generated in this section. The ScaleNet model with $\alpha = [0.5,1]$ is used for the 2-model ScaleNet. The ResNet50 architecture contains four stages after the first conv. layer, where each stage includes a convolution and identity blocks. Each conv. block and identity block consist of 3 conv. layers. Attention maps based on Grad-CAM are produced according to the first three stages of the ResNet50 to explore the results in more detail. Fig.~\ref{GRADCAM} indicates that both models learn the high-level semantic representations of the image to fulfill the rotation-prediction tasks. The attention maps of the first stage of the ScaleNet and RotNet indicate the ScaleNet catches more edge information than the RotNet, which confirms that the edge information is of vital importance for the rotation-prediction task. When the model goes more in-depth, the ScaleNet model focuses on more specific objects while the areas of interest in RotNet are distributed uniformly. Based on the attention maps in Stage-3, the ScaleNet already focuses on the head of the cat and the dog collar while the RotNet model is trying to understand the shape of the animals. The visualization result implies that the ScaleNet focuses more on the edge information and specific semantic features, which efficiently improves image classification tasks~\cite{edgeAbdillah,marmanis2018classification}. The current study results indicate that the ScaleNet learns more high-level semantic visual representations than the RotNet, which are crucial in many computer vision tasks such as object detection and segmentation~\cite{DBLP:journals/corr/rotation}.



\section{Conclusion}
An unsupervised representation learning method named ScaleNet is proposed in this study. The ScaleNet method trains multi-scale images to extract high-quality representations. The results demonstrate that ScaleNet outperforms the RotNet with different architectures such as ResNet50 and AlexNet, especially in limited information. The ScaleNet model not only learns detailed features, but the experiments outlined in this study also show that training the ScaleNet on small datasets effectively improves the performance of the RotNet on larger datasets. Furthermore, the ScaleNet process is proved to enhance the performance of other self-supervised learning models such as SimCLR. The ScaleNet is an excellent step towards improving the performance of self-supervised learning in the presence of limited information. Future studies are categorized in three steps: (i) investigating the influence of other affine transformations apart from scales, such as stretching the image. (ii) investigating the impact of additional image information apart from color and corner. (iii) implementing the ScaleNet approach in other self-supervised learning models. 
\section{Acknowledgment}
This study is supported by Google Cloud Platform (GCP) Research by providing credit supports to implement all deep learning algorithms related to SimCLR and ImageNet using virtual machines. The author would like to thank J. David Frost, Kevin Tynes, and Russell Strauss for their feedback on the draft.  
%
%
%

\bibliographystyle{splncs04}
\bibliography{refer_from_CVPR}

\end{document}